\pdfoutput=1

\documentclass[11pt]{article}

\usepackage[]{acl}
\usepackage{booktabs}
\usepackage{times}
\usepackage{latexsym}
\usepackage{multirow} 
\usepackage{eqnarray,amsmath}

\usepackage[T1]{fontenc}

\usepackage[utf8]{inputenc}

\usepackage{microtype}

\usepackage{inconsolata}

\usepackage{graphicx}
\usepackage{xcolor}

\title{Towards Prompt Generalization: \\Grammar-aware Cross-Prompt Automated Essay Scoring}

\author{First Author \\
  Affiliation / Address line 1 \\
  Affiliation / Address line 2 \\
  Affiliation / Address line 3 \\
  \texttt{email@domain} \\\And
  Second Author \\
  Affiliation / Address line 1 \\
  Affiliation / Address line 2 \\
  Affiliation / Address line 3 \\
  \texttt{email@domain} \\}

 \author{Heejin Do$^{1}$, Taehee Park$^{1}$, Sangwon Ryu$^{1}$, Gary Geunbae Lee$^{1,2}$ \\
  \centering
  \begin{tabular}[t]{c}
    $^{1}$Graduate School of Artificial Intelligence, POSTECH, Republic of Korea \\
    $^{2}$Department of Computer Science and Engineering, POSTECH, Republic of Korea \\
    \texttt{\{heejindo, taehpark, ryusangwon, gblee\}@postech.ac.kr} \\
  \end{tabular}
}

\begin{document}
\maketitle
\begin{abstract}

In automated essay scoring (AES), recent efforts have shifted toward cross-prompt settings that score essays on unseen prompts for practical applicability. However, prior methods trained with essay-score pairs of specific prompts pose challenges in obtaining prompt-generalized essay representation. In this work, we propose a grammar-aware cross-prompt trait scoring (GAPS), which internally captures prompt-independent syntactic aspects to learn generic essay representation. We acquire grammatical error-corrected information in essays via the grammar error correction technique and design the AES model to seamlessly integrate such information. By internally referring to both the corrected and the original essays, the model can focus on generic features during training. Empirical experiments validate our method's generalizability, showing remarkable improvements in prompt-independent and grammar-related traits. Furthermore, GAPS achieves notable QWK gains in the most challenging cross-prompt scenario, highlighting its strength in evaluating unseen prompts.

\end{abstract}

\section{Introduction}
Automated essay scoring (AES) emerged as a viable alternative to human graders to assist language learners in acquiring writing skills, alleviating the burden and costs of grading. To practically supply AES in educational situations, the model's capability to generalize well to new prompts (i.e., unseen in training) is essential yet challenging \cite{li-ng-2024-conundrums}. Accordingly, unlike the earlier prompt-specific AES systems, which aim to assess essays written on the seen prompts \cite{taghipour2016neural, dong2016automatic, wang2022use, do-etal-2024-autoregressive, do-etal-2024-autoregressive-multi}, recent attention increasingly moves on cross-prompt AES to grade new prompts' essays \cite{ridley2021automated, do2023prompt, chen-li-2024-plaes, li-ng-2024-conundrums}.

To achieve cross-prompt scoring with multiple trait setting, previous studies primarily focus on learning essay representation with score labels in concatenation with prompt-independent features \cite{jin2018tdnn, ridley2020prompt, li2020sednn, ridley2021automated, chen-li-2023-pmaes, do2023prompt, li-ng-2024-conundrums}. To obtain consistent essay representations, some studies additionally developed contrastive learning \cite{chen-li-2023-pmaes, chen-li-2024-plaes} or prompt-aware networks \cite{do2023prompt, jiang-etal-2022-promptbert}. However, as these models are typically trained on essays responding to specific prompts differ from the target, generalizability to unseen prompts still remains a challenge. Notably, they exhibit the lowest performance in the most prompt-agnostic \textit{Conventions} trait ($20\%$ gap to the best trait), further highlighting the shortcomings.

In this work, we propose a grammar-aware cross-prompt essay trait scoring, which integrates grammar error correction (GEC) before the scoring process. By informing the model of the syntactic errors contained in the essay, our method facilitates capturing generic syntax information during scoring. Internally, we design a shared structure to trade knowledge between original and corrected essays, facilitating accurate score derivation. As non-semantic aspects are less dependent on prompts, our grammar-aware learning via directly providing error-corrected essays leads to the intrinsic acquisition of generic essay representation.

Empirical experiments demonstrate that our grammar-aware method assists in capturing generic aspects, enhancing related trait-scoring performances in cross-prompt settings. Notably, significant enhancements observed in prompt-agnostic traits, such as \textit{Conventions} and \textit{Sentence Fluency}, support the advancement towards prompt generalized representation. Surprisingly, informing error-corrected essays also improves semantic traits, such as \textit{Content} and \textit{Narrativity}, suggesting that referring to a revised essay has the potential for facilitating a deeper contextual understanding.  




\section{Related Works}


To improve the performance of AES, several studies have auxiliary trained the model with various other tasks such as morpho-syntactic labeling, type and quality prediction, and sentiment analysis \cite{gec_aes1, gec_aes2, gec_aes3}. Instead of jointly training auxiliary tasks, our direct use of corrected text output significantly reduces the training burden.

There have been attempts to apply grammatical error information. Suggesting that detecting grammatical errors is a beneficial indicator for the quality of the essay, \citet{gec_aes5} jointly trained grammatical error detection task with the scoring model. Also, \citet{gec_aes6} utilize grammatical features proposed by \citet{criterial_feature} and \citet{gec_aes7} use GEC to measure the number of grammar corrections. Unlike the existing studies, we directly utilize the text output generated by the GEC without additional training as input for the scoring model.
%


\section{Method: GAPS}
Our method comprises two main steps: (1) Essay correction and (2) Grammar-aware Essay Scoring. Initially, we automatically identify the grammar errors included in the essay and then pass them to the scoring model along with the original essay.  

\subsection{Essay Correction}
\begin{figure}[t]
\centering
\includegraphics[width=\linewidth]{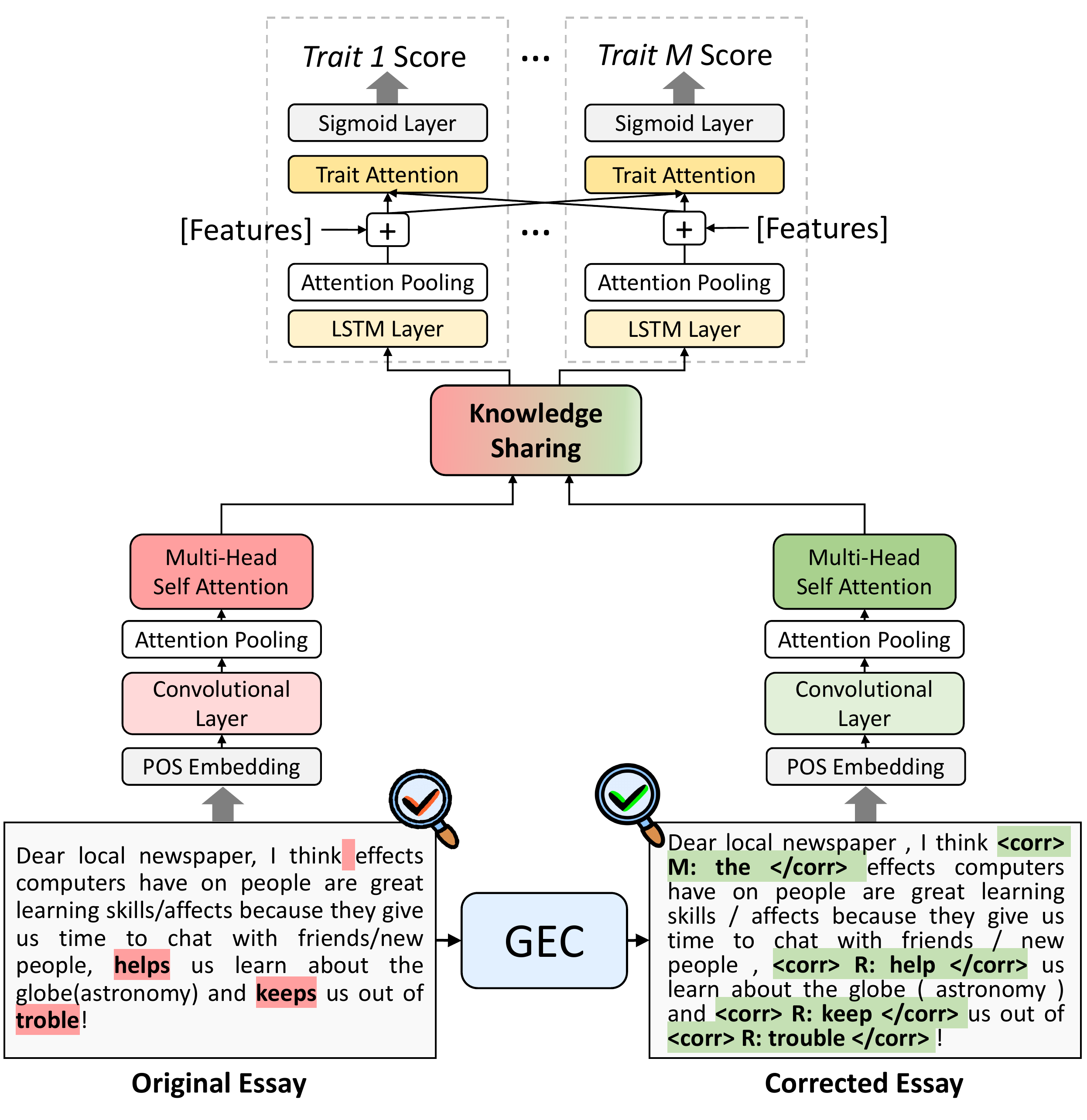}
\caption{The overview of the proposed GAPS method.}
\label{fig1}
\end{figure}

We employ the T5-based pre-trained GEC model \cite{gec_simple} to obtain the grammar-corrected essay text without additional training. The student's original essay, which contained diverse types of errors, is input into the model, and the cleaned essay is output. 
Grounded on one of the representative error types presented in the error annotation toolkit (ERRANT) \cite{errant}\footnote{\url{https://github.com/chrisjbryant/errant}}, we classify errors into three major categories: Missing (M), Replacement (R), and Unnecessary (U). Missing refers to a required token that is not present but must be inserted, replacement indicates the substituted token that is revised, and unnecessary means the deleted token that does not fit in the syntax. For the input essay, we add the correction tag, \texttt{<corr>} \texttt{Category: Token} \texttt{</corr>}, for the identified error corrections. For instance, in Figure~\ref{fig1}, for the missed token \texttt{``the"} in the essay, \texttt{<corr> M: the </corr>} is applied. Explicitly notifying the revisions could place greater emphasis during training.    

\subsection{Grammar-aware Essay Scoring}

\paragraph{Essay encoders}
We construct individual essay encoders for the original and corrected essays, respectively, but with the same structure. Our design intends to first understand each document and then share the informed knowledge. We believe enabling the model to internally distinguish each element separately, rather than combining or concatenating them, facilitates more sophisticated information exchange in subsequent layers. 

We adopt a hierarchical structure for essay encoding, which obtains trait-specific document-level representations based on sentence-level representations \cite{dong2017attention, ridley2021automated}. To obtain a generalized representation, we employ part-of-speech (POS) embedding\footnote{NLTK toolkit is used: \url{https://www.nltk.org/}}, as in previous studies \cite{ridley2021automated, do2023prompt, chen-li-2024-plaes}. After passing through POS embedding, the output $c_i$ from the 1D convolution layer \cite{kim-2014-convolutional} is subjected to attention pooling layer \cite{dong2017attention}: $\mathbf{s} = \mathrm{Pooling_{att}}([c_1:c_w])$, where $w$ denotes the number of words in the sentence. To effectively capture all parts of the essays, we adopt multi-head self-attention \cite{vaswani2017attention}, motivated by \cite{do2023prompt}:
\begin{eqnarray}
\mathrm{H_{i}}=\mathrm{att}(SW_i^{1}, SW_i^{2}, SW_i^{3}) \label{eq8}\\
\mathrm{M}=\mathrm{concat}(\mathrm{H_{1},..., H_{h}})W^{O} \label{eq9}
\end{eqnarray}
where $\mathrm{H_i}$ and $\mathrm{att}$ indicate the $i$-th head and the scaled-dot product attention, respectively. $W_i^{1\cdot 3}$ is the parameter matrices. Then, the LSTM layer \cite{hochreiter1997long} is applied, followed by the attention pooling, obtaining the original essay vector, $E_o$, and the grammar-corrected essay vector, $E_g$.

\paragraph{Knowledge-sharing layers}
Given two representations of the original and the grammar-corrected essay, we introduce the knowledge-sharing layers via the cross-attention leveraging multi-head attention mechanism. Specifically, with the original essay vector $E_o$ as the key and value and the grammar-corrected vector $E_g$ as the query, the knowledge-sharing layer is defined as follows: 
\begin{eqnarray}
\mathrm{H_{i}}=\mathrm{att}(E_gW_i^{1}, E_{o}W_i^{2}, E_{o}W_i^{3}) \label{eq3}\\
\mathrm{M}=\mathrm{concat}(\mathrm{H_{1},..., H_{h}})W^{O} \label{eq4}
\end{eqnarray}
Subsequently, $m$ trait-specific layers are obtained for $m$ distinct traits. Following previous studies, we concatenate the prompt-independent features of \citet{ridley2021automated} to each trait-wise essay representation vector. To refer to other traits' representations during training, we employ the trait-attention mechanism \cite{ridley2021automated}. 

\paragraph{Training} For the loss function, we use mean squared error: $
    \mathrm{L}(y,\hat{y}) = \frac{1}{\mathrm{n\cdot m}}\sum_{i=1}^{\mathrm{n}}\sum_{j=1}^{\mathrm{m}}(\hat{y}_{ij}-y_{ij})^2 $, with $n$ number of essays and $m$ trait scores. As different prompts are evaluated by different traits (Apendix~\ref{tab: dataset statistics}), the masking mechanism is applied to mark empty traits as 0 \cite{ridley2021automated}.

\section{Experiments}
For experiments, we use the Automated Student Assessment Prize (ASAP\footnote{\url{https://www.kaggle.com/c/asap-aes}}) and ASAP++\footnote{\url{https://lwsam.github.io/ASAP++/lrec2018.html}} \citep{mathias2018asap++} dataset, which are publicly available and representative for AES. The dataset includes eight prompts and corresponding essays written in English, and multiple trait scores are assigned by human raters (Table~\ref{tab: dataset statistics}). 

In the cross-prompt setting, each target prompt is used for testing, while the other seven prompts are used for training. For instance, when the target prompt is 8, only 1--7 prompts are used in training and testing with P8. We used the 2080ti GPU, batch 10, epoch 50, selecting the model with the best validation. 
We use the efficient GEC model proposed by \citet{gec_simple}, which is pre-trained on sentence-level corrupted mC4 corpus and fine-tuned on cLang-8\footnote{\url{https://github.com/google-research-datasets/clang8}} dataset. As the official code is absent, we used an open-source implementation of the model\footnote{\url{https://github.com/gotutiyan/gec-t5}}, fine-tuned on English benchmark data on T5 \cite{2020t5}, achieving F0.5 scores of 65.01 on CoNLL-2014-test \cite{ng-etal-2014-conll} and 70.32 on BEA-19 test set \cite{bryant-etal-2019-bea}.

\begin{table}
\centering
\scalebox{
0.67}{
\begin{tabular}{cccccccc}
\toprule
{\textbf{Pr}} & {\textbf{Evaluated Traits}} & {\textbf{\# Essays}}  & {\textbf{Essay Type}}\\
\hline
P1 & Cont, Org, WC, SF, Conv & 1,783 & Argumentative \\
P2 & Cont, Org, WC, SF, Conv & 1,800 & Argumentative   \\
P3 & Cont, PA, Lan, Nar & 1,726 & Source-Dependent   \\
P4 & Cont, PA, Lan, Nar & 1,772 & Source-Dependent   \\
P5 & Cont, PA, Lan, Nar  & 1,805 & Source-Dependent \\
P6 & Cont, PA, Lan, Nar & 1,800 & Source-Dependent  \\
P7 & Cont, Org, Conv, Style & 1,569 & Narrative  \\
P8 & Cont, Org, WC, SF, Conv, Voice & 723 & Narrative  \\
\bottomrule
\end{tabular}}
\caption{ASAP/ASAP++ combined dataset statistics. \textit{Pr} denotes the prompt number. WC: \textit{Word Choice}; PA: \textit{Prompt Adherence}; Nar: \textit{Narrativity}; Org: \textit{Organization}; SF: \textit{Sentence Fluency}; Conv: \textit{Conventions}; Lang: \textit{Language}.}
\label{tab: dataset statistics}
\end{table}

\definecolor{sky}{RGB}{204, 229, 255}
\definecolor{pin}{RGB}{255,204,229}

\begin{table*}[t]
\centering

\scalebox{
0.7}{
\begin{tabular}{l|ccccccccc|c|c}
\hline
\multirow{2}{*}{\textbf{Model}} & \multicolumn{9}{c|}{\textbf{Traits}} & \multirow{2}{*}{\textbf{AVG}} & \multirow{2}{*}{\textbf{SD(↓)}} \\ \cmidrule(rl){2-10}
 & Overall & Content & Org & WC & SF & Conv & PA & Lang & Nar &  & -\\
\hline
Hi att   & 0.453   & 0.348 & 0.243 & 0.416 & 0.428 & 0.244 & 0.309 & 0.293 & 0.379 & 0.346 & -  \\
AES aug   & 0.402   & 0.342 & 0.256 & 0.402 & 0.432 & 0.239 & 0.331 & 0.313 & 0.377 & 0.344 & - \\
PAES \citep{ridley2020prompt}& 0.657 & 0.539 & 0.414 & 0.531 & 0.536 & 0.357 & 0.570 & 0.531 & 0.605 & 0.527 & -\\
CTS \citep{ridley2021automated} & 0.670 & 0.555 & 0.458 & 0.557 & 0.545 & 0.412 & 0.565 & 0.536 & 0.608 & 0.545 & -\\
PMAES \cite{chen-li-2023-pmaes}   & 0.671   & 0.567 & 0.481 & 0.584 & 0.582 & 0.421 & {0.584} & 0.545 & 0.614 & 0.561 & -  \\
PLAES \cite{chen-li-2024-plaes} & {0.673} & {0.574} & {0.491} & 0.579 & 0.580 & 0.447 & 0.601 & 0.554 & {0.631} & {0.570} & - \\
ProTACT \textbf{[{TA}+\colorbox{sky!80}{PA}]} \cite{do2023prompt} & \textbf{0.674} & 0.596 & \textbf{0.518} & \textbf{0.599} & 0.585 & 0.450 & {0.619} & 0.596 & 0.639 & 0.586 & ±0.009\\
\midrule
Single Encoder  & {0.673} & 0.567 & 0.480 & 0.578 & 0.573 & 0.437 & 0.571 & 0.548 & 0.612 & 0.560 & ±0.012 \\
\textbf{GAPS [\colorbox{pin!70}{GA}]} & 0.672 & {0.573} & {0.485} & {0.580} & {0.586} & {0.451} & {0.582} & {0.567} & {0.630} & {0.570} & ±0.014\\
\midrule
\textbf{GAPS [{TA}+\colorbox{pin!70}{GA}]} & 0.669 & 0.595 & 0.514 & 0.585 & 0.579 & 0.465 & 0.615 & 0.603 & 0.648 & 0.586   & ±0.017\\
\textbf{GAPS [{TA}+\colorbox{sky!80}{PA}+\colorbox{pin!70}{GA}]} & 0.670  & \textbf{0.597}  & {0.515}  & 0.595  & \textbf{0.590} & \textbf{0.472}  & \textbf{0.621}  & \textbf{0.608}  & \textbf{0.650}  & \textbf{0.591}  & ±0.011\\
\hline
\end{tabular}}
\caption{\label{tab1}
Five runs averaged QWK scores over all prompts for each \textbf{trait}. {TA} and \colorbox{sky!80}{PA} denote the prompt-aware and trait-aware methods in ProTACT \cite{do2023prompt}, respectively, while \colorbox{pin!70}{GA} represents our grammar-aware approach.}

\smallskip
\smallskip
\smallskip

\scalebox{
0.7}{
\begin{tabular}{l|cccccccc|c|c}
\toprule
\multirow{2}{*}{\textbf{Model}} & \multicolumn{8}{c|}{\textbf{Prompts}} & \multirow{2}{*}{\textbf{AVG}} & \multirow{2}{*}{\textbf{SD(↓)}} \\ \cmidrule(rl){2-9}
 & 1 & 2 & 3 & 4 & 5 & 6 & 7 & 8 &  &  \\
\hline
Hi att   & 0.315 & 0.478 & 0.317 & 0.478 & 0.375 & 0.357 & 0.205 & 0.265 & 0.349  & -  \\
AES aug  & 0.330 & 0.518 & 0.299 & 0.477 & 0.341 & 0.399 & 0.162 & 0.200 & 0.341 & - \\
PAES \citep{ridley2020prompt} & 0.605 & 0.522 & 0.575 & 0.606 & {0.634} & 0.545 & 0.356 & 0.447 & 0.536 & -\\
CTS \citep{ridley2021automated}& 0.623 & 0.540 & 0.592 & 0.623 & 0.613 & 0.548 & 0.384 & 0.504 & 0.553 & -\\
PMAES \cite{chen-li-2023-pmaes}  & {0.656} & 0.553 & 0.598 & 0.606 & 0.626 & 0.572 & 0.386 & 0.530 & 0.566 & - \\
PLAES \cite{chen-li-2024-plaes} & 0.648 & 0.563 & 0.604 & 0.623 & {0.634} & \textbf{0.593} & 0.403 & 0.533 & {0.575} & -  \\
ProTACT \textbf{[{TA}+\colorbox{sky!80}{PA}]} & 0.647 & 0.587 & 0.623 & 0.632 & \textbf{0.674} & 0.584 & 0.446 & \textbf{0.541} & 0.592 & ±0.016\\
\midrule
Single Encoder & 0.633 & 0.562 & 0.595 & 0.620 & 0.616 & 0.562 & 0.406 & {0.534} & 0.566 & ±0.016\\
\textbf{GAPS [\colorbox{pin!70}{GA}]} & 0.631 & {0.587} & {0.610} & {0.637} & 0.614 & 0.580 & {0.421} & 0.520 &{0.575} & ±0.016\\
\hline
\textbf{GAPS [TA+\colorbox{pin!70}{GA}]}  & 0.627 & \textbf{0.626} & 0.633 & 0.640 & 0.660 & {0.591} & \textbf{0.469} & 0.494 & 0.593 & ±0.022\\
\textbf{GAPS [TA+\colorbox{sky!80}{PA}+\colorbox{pin!70}{GA}]}  & \textbf{0.654} & 0.614 & \textbf{0.636} & \textbf{0.646} & 0.665 & 0.590 & \textbf{0.469} & 0.498 & \textbf{0.597} & ±0.019\\
\hline
\end{tabular}}
\caption{\label{tab2}
Five runs averaged QWK scores over all traits for each \textbf{prompt}; \textit{SD} is the averaged standard deviation for five seeds, and \textbf{bold} text indicates the highest value.}

\end{table*}




\section{Results and Discussions}

\paragraph{Comparison with single encoder} As our goal is to validate the impact of the proposed grammar-aware approach, we primarily compare GAPS against the \textit{Single Encoder} model, which processes only the original essay without incorporating grammatical error-corrected versions yet within our designed structure. Trait-wise results in Table~\ref{tab1} highlight that referring to corrected essays with GAPS remarkably enhances the scoring performance for all traits except for \textit{Overall}. Notably, the improvement is more pronounced in syntactic and lexical-related traits; nevertheless, the observed QWK enhancements in contextual assessment traits suggest that our method also facilitates the capture of semantic aspects. Prompt-wise results in Table~\ref{tab2} demonstrate that GAPS consistently outperforms the single-encoder model, confirming the efficacy of referring to error correction information.


\paragraph{Generalizability across prompts}
Grammar serves as a universal, prompt-agnostic criterion for evaluation, largely unaffected by the specific instructions within a prompt; thus, it can be a great indicator for prompt generalization. This is particularly evident in the \textit{Convention} trait, which evaluates writing conventions such as spelling and punctuation, independent of prompt-relevant information \cite{mathias2018asap++}. While even robust previous models, such as PMAES \cite{chen-li-2023-pmaes}, PLAES \cite{chen-li-2024-plaes}, and ProTACT \cite{do2023prompt}, have shown significantly lower performance in this trait, GAPS demonstrates substantial improvements in the \textit{Convention}. These results emphasize the robustness of our method's prompt generalization capabilities. Furthermore, in a prompt-wise examination, we observe substantial performance gains for the challenging Prompt 7, which presents a difficult cross-prompt setting due to its differences in type and evaluated trait compositions (Table~\ref{tab: dataset statistics}). Although Prompt 8 shares the same type, it is constrained by a smaller dataset of only 723 samples. Thus, the notable improvements in Prompt 7 indicate GAPS' ability to effectively evaluate essays of new, unseen prompts, even in more challenging settings.


\begin{figure}
\centering
\includegraphics[width=0.95\linewidth]{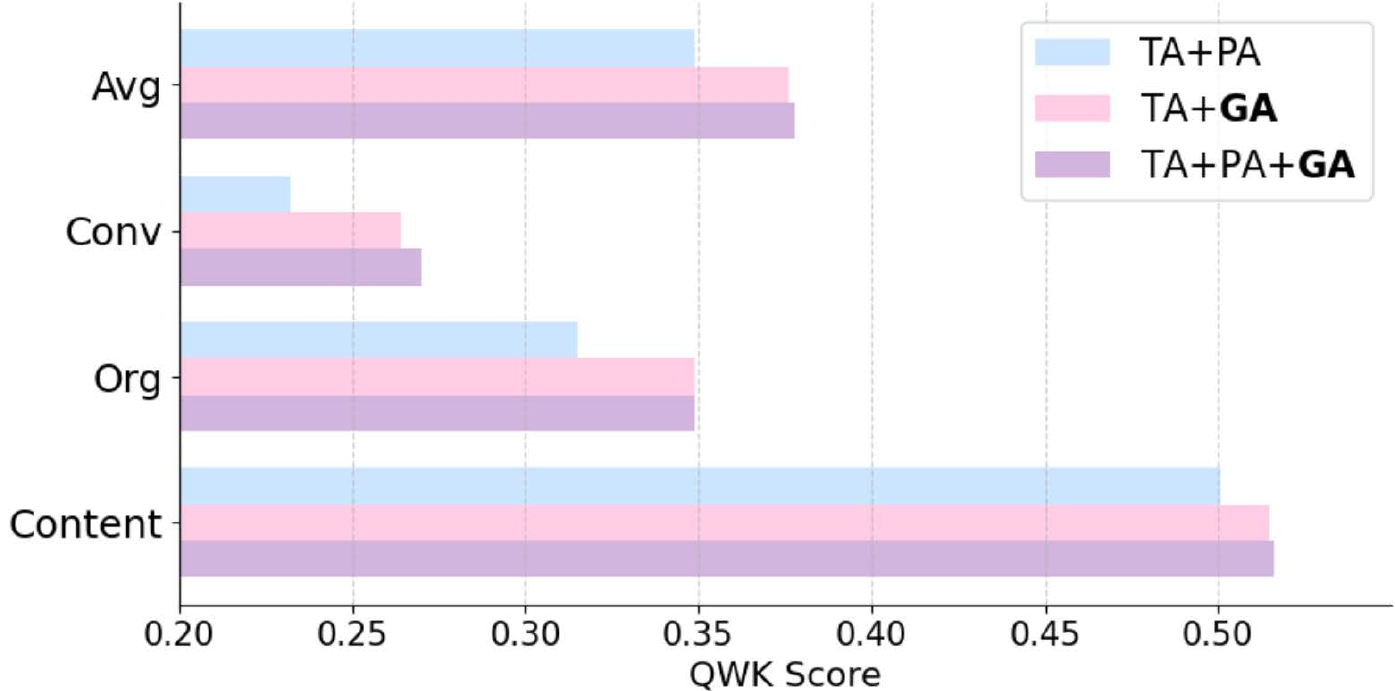}
\caption{QWK scores for traits evaluated in Prompt 7.}
\label{fig2}
\end{figure}

\begin{table*}[t]
\centering
\scalebox{
0.7}{
\begin{tabular}{l|ccccccccc|c|c}
\hline
\multirow{2}{*}{\textbf{Model}} & \multicolumn{9}{c|}{\textbf{Traits}} & \multirow{2}{*}{\textbf{AVG}} & \multirow{2}{*}{\textbf{SD(↓)}} \\ \cmidrule(rl){2-10}
 & Overall & Content & Org & WC & SF & Conv & PA & Lang & Nar &  & \\
\hline

GAPS & \textbf{0.672} & \textbf{0.573} & 0.485 & \textbf{0.580} & \textbf{0.586} & \textbf{0.451} & \textbf{0.582} & 0.567 & 0.630 & \textbf{0.570} & ±0.014 \\
w/o KS & 0.672 & 0.570 & \textbf{0.488} & 0.561 & 0.564 & 0.446 & 0.570 & \textbf{0.571} & \textbf{0.632} & 0.564 & ±0.015\\
w/o GCT &  0.667 & 0.559 & 0.467 & \textbf{0.580} & 0.569 & 0.420 & 0.571 & 0.546 & 0.611 & 0.554	& ±0.011\\
\hline
\end{tabular}}
\caption{\label{tab3}
Five runs averaged ablation QWK results over all prompts for each \textbf{trait}. KS and GCT denote the \textit{Knowledge Sharing} and \textit{Grammar Correction Tagging}, respectively. }
\end{table*}

\paragraph{Impact of grammar-aware vs. prompt-aware approaches}
We directly compare GAPS, our grammar-aware (GA) approach, with ProTACT’s prompt-aware (PA) method, which leverages prompt information directly. Since ProTACT also introduces trait-relation-aware (TA) methods, such as trait-similarity loss, we incorporate TA into our model for a fair comparison (i.e., TA+GA vs. TA+PA). Results in Table~\ref{tab1} show that PA excels in \textit{Organization}, \textit{Word Choice}, and \textit{Sentence Fluency}, indicating its strength in capturing logical flow and prompt adherence. In contrast, GA outperforms PA in \textit{Conventions}, \textit{Language}, and \textit{Narrativity}, demonstrating its superiority in enhancing grammatical correctness and structural coherence. Notably, GA’s impact on \textit{Conventions} emphasizes the direct benefits of referring to grammatically corrected contexts. For most traits, using GA with PA yields better performance.

We also investigate the effects of GA on the traits evaluated in Prompt 7 (Figure~\ref{fig2}). Interestingly, in this low-resource cross-prompt scenario, where similar types are scarce, GA outperforms PA in all traits. This result suggests that incorporating grammar-revised essays is much more beneficial than relying on prompt information alone, especially in challenging cross-prompt settings.



\paragraph{Effects of knowledge sharing}
We further examine the impact of the designed knowledge-sharing layer by comparing GAPS with a version that omits the knowledge-sharing component (Table~\ref{tab3}; w/o KS). Instead of using Equations~\ref{eq3} and ~\ref{eq4}, we simply concatenate the obtained $E_o$ and $E_g$ vectors and subsequently input them to the LSTM layer. Removing the KS module results in a marked decline in the \textit{Word Choice}, \textit{Sentence Fluency}, and \textit{Convention} traits, underscoring the pivotal role of knowledge sharing in effectively capturing both structural and syntactic features. Without the KS module, the model struggles to integrate the original and grammar-corrected essay representations, which hinders its ability to make accurate judgments of these traits.

\paragraph{Effects of grammar correction tagging}
To investigate whether the inclusion of grammar correction tags is effective, we conducted an ablation study to eliminate the tags, utilizing only the pure corrected essay (Table~\ref{tab3}). Notably, specifying the correction tags in the essay significantly improves scoring performance across most traits, revealing the importance of key entity identification for balanced generalization. These findings are consistent with existing studies, which show that underscoring the key entities improves performance on downstream tasks \cite{ryu24_interspeech}.     

\section{Conclusion}
We propose a grammar-aware cross-prompt trait scoring to enhance prompt generalizability. By directly utilizing grammar error-corrected essays as the input, the model can learn more syntactic-aware representations of essays. In addition, we introduce tagging the corrected tokens, which leads the model to better focus on critical parts for grading. Our experiments demonstrate that grammar-aware essay representation obtained with our straightforward model structure remarkably assists the scoring of lexical or grammatical traits. Further, the notable performance increase in the most challenging prompt implies our model's internal acquisition of prompt-independent features.

\section{Limitations}
We have explored the use of grammar error correction to assist in obtaining invariant essay representation for cross-prompt trait scoring. Our limitation relates to the possible dependency on the GEC performance, which is not handled in this work. Although we used the robust and effective GEC method, further experiments with different models will provide another room for scoring quality improvement on cross-prompt settings.   

\section{Ethical Statement}
We used publicly available datasets in this work.

\section*{Acknowledgments}
This research was partially supported by the MSIT (Ministry of Science and ICT), Korea, under the ITRC (Information Technology Research Center) support program (IITP-2025-2020-0-01789) supervised by the IITP (Institute for Information \& Communications Technology Planning \& Evaluation) (50\%). It was also supported by an IITP grant funded by the Korea government (MSIT) (No.RS-2019-II191906, Artificial Intelligence Graduate School Program (POSTECH)) (50\%).

\bibliography{main}

\end{document}